\documentclass[12pt]{article}
\usepackage[english]{babel}
\usepackage{amsmath,amsthm}
\usepackage{amsfonts}
\usepackage{graphicx}
\usepackage{epstopdf}

\theoremstyle{definition}

\theoremstyle{remark}

\numberwithin{equation}{section}

\def\beq{\begin{equation}}
\def\eeq{\end{equation}}
\def\cF{\mathcal{F}}
\def\cH{\mathcal{H}}
\def\cO{\mathcal{O}}
\def\beps{{\boldsymbol{\varepsilon}}}
\def\bc{\mathbf{c}}
\def\bs{\mathbf{s}}
\def\bp{\mathbf{p}}
\def\bg{\mathbf{g}}
\def\bh{\mathbf{h}}
\def\bI{\mathbf{I}}
\def\bQ{\mathbf{Q}}
\def\bD{\mathbf{D}}
\def\bS{\mathbf{S}}
\begin{document}

\title{Fast and numerically stable circle fit}
\author{H.~Abdul-Rahman$^1$ and N.~Chernov$^1$}

\date{}

\maketitle


\begin{abstract}
We develop a new algorithm for fitting circles that does not have
drawbacks commonly found in existing circle fits. Our fit achieves
ultimate accuracy (to machine precision), avoids divergence, and is
numerically stable even when fitting circles get arbitrary large.
Lastly, our algorithm takes less than 10 iterations to converge, on
average.
\end{abstract}


\centerline{Keywords: fitting circles, geometric fit,
Levenberg-Marquardt, Gauss-Newton}

\footnotetext[1]{Department of Mathematics, University of Alabama at
Birmingham, Birmingham, AL 35294, USA; Email:$\ $houssam@uab.edu,
chernov@uab.edu}

\section{Introduction}

Fitting circles and circular arcs to observed points is a basic task in
pattern recognition and computer vision
\cite{Ahnbook,AW04,Cbook,CL1,Cr83,GGS94,Jo94,Ka91,Sp96}. Some authors
assert that ``\textit{most of the objects in the world are made up of
circular arcs and straight lines}''; see \cite{Pe78,YF96}.

The classical least squares fit minimizes geometric distances from the
observed points to the fitting circle:
\begin{equation} \label{main}
    F(a,b,R)=\sum_{i=1}^{n}\bigl[\sqrt{(x_i-a)^2+(y_i-b)^2}-R\bigr]^2
    \ \ \to\ \ \min
\end{equation}
Here $(x_i,y_i)$ denotes the observed points, $(a,b)$ the center and
$R$ the radius of the fitting circle.

The geometric fit \eqref{main} has many attractive features. It is
invariant under translations, rotations, and scaling, i.e., the best
fitting arc does not depend on the choice of the coordinate system. It
provides the maximum likelihood estimate under standard statistical
assumptions \cite{Ch65,CL2,Cbook}. The minimization of geometric
distances is often regarded as the most desirable solution\footnote{In
particular, it has been prescribed by a recently ratified standard for
testing the data processing software for coordinate metrology
\cite{Ahnbook}.} of the fitting problem, albeit hard to compute in some
cases.

Our paper is devoted to practical algorithms for minimization of
\eqref{main}. Theoretical aspects of the circle fitting problem are
covered elsewhere: for the existence and uniqueness of the global
minimum of the objective function $F(a,b,R)$ see
\cite{CL2,Cbook,Ni02,ZC06a}, for its differentiability see Lemma~7 in
\cite{Ni04a}, etc.

Most authors solve \eqref{main} by general minimization algorithms,
such as Gauss-Newton (GN) \cite{GGS94,Jo94} or Levenberg-Marquardt (LM)
\cite{CL1,Sh98}; see a review \cite{Cbook}. Circle-specific schemes
exist \cite{La87,Sp96} but they converge linearly and often take
hundreds of iterations \cite{Cbook,CL1}, which makes them impractical.

The GN and LM normally converge in 5-10 iterations, but they have a
number of known issues. First, they occasionally diverge. It was shown
\cite{CL1} (see a detailed proof in \cite[Section~3.8]{Cbook}) that
there is a valley in the parameter space that extends to infinity,
along which the objective function slowly decreases but remains above
its minimum value. Thus if the minimization algorithm starts in that
valley (or gets there by chance) it will be forced to move away from
the minimum of \eqref{main}.

Numerical tests \cite{CL1} show that if the initial guess is picked at
random, the chance of divergence may be as high as 50\%. If the initial
guess is supplied by an algebraic circle fit (such as K{\aa}sa fit
\cite{Ka76}), then the chances of divergence are very low, but it is
still possible (see an example in \cite[Section~5.13]{Cbook}).

One way to avoid divergence is to use the so-called algebraic
parameters, in which the circle equation is $A(x^2+y^2) + Bx + Cy + D =
0$ with additional constraint $B^2+C^2=4AD+1$; see details in
\cite{Cbook,CL1}. One can determine $A,B,C,D$ for the best fitting
circle and then convert them to $a,b,R$. Such an algorithm was proposed
in \cite{CL1}, and it indeed converges to a minimum of \eqref{main}
from any starting point. But using algebraic parameters $A,B,C,D$ leads
to very complicated formulas, and the resulting algorithm is about 10
times slower than the one using the geometric parameters $a,b,R$ (see
\cite{Cbook,CL1}). There is also a possible loss of accuracy at the
stage of converting the algebraic parameters to the geometric ones. Our
numerical tests (Section~\ref{Numerical results}) demonstrate these
drawbacks.

The second issue is accuracy. Standard algorithms use an adaptive step
procedure: if the value of the objective function \eqref{main} is
\emph{not} decreasing at the next iteration, the latter is rejected and
the step is recomputed by adjusting the value of a control parameter.
This `acceptance rule' makes the accuracy of the parameter estimates no
better than $\cO(\beps^{1/2})$, where $\beps$ denotes the machine
precision ($\beps \approx 2\cdot 10^{-16}$ in the standard double
precision arithmetic); we explain this below. In fact, most algorithms
stop iterations whenever the step gets smaller than $\beps^{1/2}$; see
\cite{GGS94,NR}.

We modify this `acceptance rule' so that the accuracy of the
parameter estimates becomes $\cO(\beps)$, rather than
$\cO(\beps^{1/2})$. This makes our algorithm \emph{numerically
stable}, in a formal sense \cite{TBbook}. This ultimate accuracy is
achieved by using the \emph{norm of the gradient} of the objective
function, rather than the function itself, as the `acceptance
criterion'; see below.

One may wonder how many additional iterations it takes to reach this
ultimate accuracy. The standard Gauss-Newton and Levenberg-Marquardt
schemes are known to converge linearly in the vicinity of a minimum, so
they might take 10-15 extra iterations. Instead, we employ a version of
a `full Newton method' \cite{NR} which guarantees quadratic
convergence. So it takes just 1-2 extra iterations to reach the
ultimate accuracy.

The third issue is related to large circles. If the data points lie
along a circular arc with low curvature, then the best fitting circle
has a large radius $R$ and a far away center $(a,b)$. This leads to
catastrophic cancelation in the calculation of the function
\eqref{main} and its derivatives, so the resulting estimates get poor.
As a remedy, one can use algebraic parameters \cite{CL1} or other
special parameters \cite{Ka91}, but again this leads to very
complicated formulas and involves an inevitable loss of accuracy at the
stage of conversion from one set of parameters to another.

We resolve this issue while staying with the natural geometric
parameters $(a,b,R)$ at all times. For large circles we use different
formulas for the objective function \eqref{main} and its derivatives,
which are mathematically equivalent to the standard formulas but are
organized differently to avoid catastrophic cancelation; see
Section~\ref{NewFormulas}.

Our numerical tests show that the resulting algorithm has the following
features: convergence to a local minimum of \eqref{main} from nearly
any initial guess (in fact, it never failed to converge in our tests),
the final accuracy $\cO(\beps)$, the average number of iterations is
only 7-8, and the average execution time is only 50\% higher compared
to standard GN and LM algorithms (which have issues as listed above).
We tested all the other published algorithms and none of them came
close to these characteristics; see Section~\ref{Numerical results}.
While the numerical tests may not `prove' the superiority of our
approach, we believe that our theoretical resolution of the above
difficult issues constitutes the real novelty of our work.

The paper is organizes as follows. We begin with a standard
modification and reduction of the objective function. In
section~\ref{Cherckpoints} we propose a ``two phase'' acceptance rule
to increase the accuracy from $\cO(\beps^{1/2})$ to $\cO(\beps)$. In
section~\ref{Full Newton} we present our version of the full Newton
method with Levenberg-Marquardt correction. In
section~\ref{NewFormulas} we derive formulas for the objective
function, its derivative and Hessian, that avoid catastrophic
cancelation for large circles. In section~\ref{escape valley} we
describe a method that prevents divergence. In Section~\ref{Numerical
results} we present our numerical tests.

\section{Reducing and Modifying the Problem}\label{Reducing}

Expression (\ref{main}) can be reduced by eliminating the radius $R$,
as the right hand side of (\ref{main}) is a quadratic polynomial in
$R$. Setting $\partial F/\partial R=0$ and solving for $R$ yields
\beq  \label{R}
   R=\overline{r}\qquad\text{where}\qquad
   r_i=\sqrt{(x_i-a)^2+(y_i-b)^2}
\eeq
Here we use the ``sample mean'' notation $\overline{r}=\frac 1n\,
\sum_{i=1}^n r_i$. Similar notation is used below for
$\overline{x}=\frac 1n\,\sum_{i=1}^{n}x_i$, $\overline{xy}=\frac
1n\,\sum_{i=1}^n x_i y_i$, etc. Now (\ref{main}) reads
\begin{align}
  F(a,b) &= \sum_{i=1}^n(r_i-\overline{r})^2 \nonumber\\
         &= \sum_{i=1}^n (x_{i}^2+y_{i}^2)+n(a^2+b^2-\overline{r}^2)-2an\overline{x}-2bn\overline{y}
\end{align}
Since the term $\sum_{i=1}^n (x_{i}^2+y_{i}^2)$  is constant, we can
drop it, then divide $F$ by $n$, and proceed to minimize the
``reduced'' objective function
\beq \label{F1}
    \mathcal{F}(a,b)=a^2+b^2-\overline{r}^2-2a\overline{x}-2b\overline{y}
\eeq
Centering the data prior to the fit is known to help reduce round-off
errors \cite{Cr83,Ta91}. This is done by translation $x_i'= x_i -
\overline{x}$ and $y_i'= y_i - \overline{y}$. It also helps to scale
the data to make their values of order one. This is done by $x_i''=
x_i'/S$ and $y_i''= y_i'/S$ where
$S=[\overline{x'x'}+\overline{y'y'}]^{1/2}$.

After one estimates the circle center $(a,b)$ using the centered and
scaled data points, one needs to rescale and retranslate it by $a
\mapsto Sa+ \overline{x}$ and $b\mapsto Sb + \overline{y}$ and then
compute $R$ by \eqref{R}.

We will assume that the data is already centered and scaled. In
particular, this makes $\overline{x}=\overline{y}=0$, hence \eqref{F1}
further reduces to
\begin{equation}\label{F1a}
    \mathcal{F}(a,b)=a^2+b^2-\overline{r}^2
\end{equation}

\section{Using a Gradient Based Acceptance Rule}\label{Cherckpoints}

\begin{figure}
\begin{center}
  \includegraphics[width=4in]{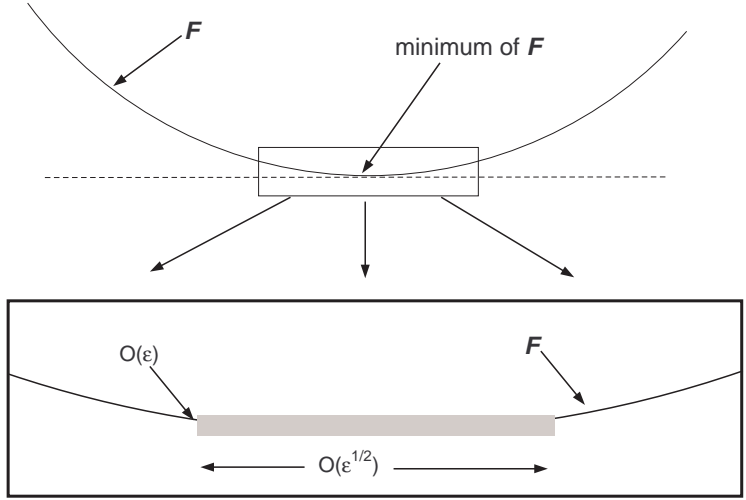}\\
  \caption{$\mathcal{F}$ has quadratic behavior in the vicinity of a minimum}\label{Checkpoint1}
\end{center}
\end{figure}

The objective function $\mathcal{F}$ is smooth, hence its gradient
vanishes at any local minimum $\bp_{\min}$ (we use notation
$\bp=(a,b)$), thus
$$
   \mathcal{F}(\bp_{\min}+\bh)-\mathcal{F}(\bp_{\min})= O(\|\textbf{h}\|^2)
$$
This means that if $\|\bh\| \leq \beps^{1/2}$, then round-off errors
make it impossible to reliably compare the values of
$\mathcal{F}(\bp_{\min}+\bh)$ and $\mathcal{F}(\bp_{\min})$ (see
Fig.~\ref{Checkpoint1}). Thus, standard minimization schemes with
adaptive step (GN, LM, or Trust Region \cite{Cbook}), which accept
the next iteration if $\cF(\bp_{\rm next}) < \cF(\bp_{\rm
current})$, are doomed to stall in the $O(\beps^{1/2})$ neighborhood
of $\bp_{\min}$. We use this acceptance rule (we call it AR1) only
outside that neighborhood.

\begin{figure}[htb]
\begin{center}
  \includegraphics[width=3.5 in]{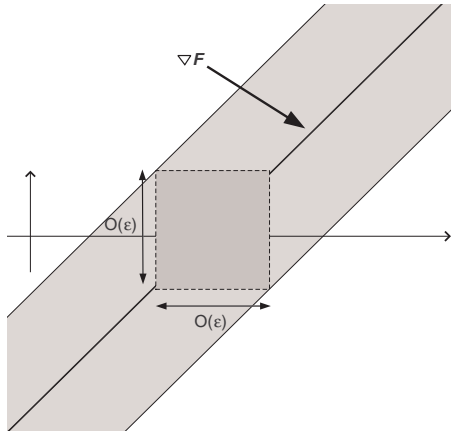}
\vspace{-30pt}
  \caption{$\nabla \mathcal{F}$ has linear behavior in the vicinity of the minimum}\label{Checkpoint2}
\end{center}
\end{figure}

Inside the $O(\beps^{1/2})$ neighborhood of $\bp_{\min}$ we use the
norm of the gradient $\nabla\cF$; i.e., we accept the next iteration
provided $\|\nabla\cF(\bp_{\rm next})\| < \|\nabla\cF(\bp_{\rm
current})\|$ (we call this rule AR2). Since $\cF$ is twice
differentiable, we have
$$
  \nabla \mathcal{F}(\bp_{\min})=0
  \quad\text{and}\quad
  \|\nabla \mathcal{F}(\bp_{\min}+\bh)\|=
  O(\|\bh\|)
$$
thus the numerically computed value of $\|\nabla \mathcal{F}(\bp)\|$
remains significantly different from zero all the way down to $\|\bh\|
= O(\beps)$; see Fig.~\ref{Checkpoint2}.

More precisely, we abandon AR1 and start using AR2 once
\begin{equation}\label{checkpoint Phase2}
    \|\nabla \mathcal{F}\|\leq \varepsilon_{\ast}\qquad
    (\varepsilon_{\ast}\sim\beps^{1/2})
\end{equation}
where $\varepsilon_{\ast}=3\times 10^{-8}$ in double precision and
$10^{-4}$ in single precision \cite{NR}.

\section{Full Newton Minimization}\label{Full Newton}

Fast (quadratic) convergence to a minimum of $\cF$ in its
$O(\beps^{1/2})$ vicinity requires the use of the full Newton method,
i.e., exact formulas for the gradient and Hessian matrix of $\cF$. By
standard formulas \cite{Cbook}, the gradient is
\begin{equation}\label{Gradient}
    \tfrac{1}{2}\,\nabla\mathcal{F}=\left[
                           \begin{array}{c}
                             a+\overline{u}\, \overline{r} \\
                             b+\overline{v}\, \overline{r} \\
                           \end{array}
                         \right]
\end{equation}
and the Hessian matrix
\begin{equation}\label{Hessian}
    \tfrac{1}{2}\,\mathcal{H}=\left[
                            \begin{array}{cc}
                              1-\overline{u}^2-\overline{r}\,\overline{vv/r} & -\overline{u}\,
                              \overline{v}+\overline{r}\,\overline{uv/r} \\
                              -\overline{u}\,
                              \overline{v}+\overline{r}\,\overline{uv/r} &
                              1-\overline{v}^2-\overline{r}\,\overline{uu/r} \\
                            \end{array}
                          \right]
\end{equation}
where we used our ``sample mean'' notation and denoted
\begin{align*}
  u_i = -\frac{\partial r_i}{\partial a}   = \frac{x_i-a}{r_i},
  \qquad
  v_i = -\frac{\partial r_i}{\partial b}   = \frac{y_i-b}{r_i}
\end{align*}
In particular, $\overline{uv/r} = \frac 1n\, \sum u_iv_i/r_i$, etc.

Our adaptive step rule is based on the Levenberg-Marquardt type
correction to the Hessian matrix:
$$
    \mathcal{H}_{\lambda}=\mathcal{H}+\lambda \bI
$$
where $\lambda>0$ is a control parameter and $\bI$ is the $2\times 2$
identity matrix. The next iteration step is then computed by
\begin{equation}\label{StepLEv}
    \bh=-\mathcal{H}_{\lambda}^{-1}\cdot(\nabla\mathcal{F})
\end{equation}
For better accuracy, we use an eigenvalue decomposition of $\cH$
\beq  \label{eigenH}
   \cH = \bQ\bD\bQ^T
\eeq
where $\bQ$ is an orthogonal matrix and $\bD = {\rm diag}\{ d_1, d_2\}$
is a diagonal matrix containing the eigenvalues of $\cH$. Since $\cH$
is a $2\times 2$ matrix, its eigendecomposition can be computed by fast
direct formulas (without using matrix algebra functions). Now
\beq
   \cH_{\lambda} = \bQ\bD_{\lambda}\bQ^T
\eeq
where
\beq
    \bD_{\lambda} = {\rm diag}\{ d_1+\lambda, d_2+\lambda\}
\eeq
and \eqref{StepLEv} takes form
\begin{equation}\label{StepLEv1}
    \bh=-\bQ\bD_{\lambda}^{-1}\bQ^T\nabla\mathcal{F}
\end{equation}
where $\bD_{\lambda}^{-1} = {\rm diag}\{ 1/(d_1+\lambda),
1/(d_2+\lambda)\}$.

We choose $\lambda$ so that (i) the matrix $\cH_{\lambda}$ is positive
definite and, moreover, (ii) the step $\bh$ is not too large. The
latter means
\beq  \label{bhalpha}
     \|\bh\| \leq h_{\max} = \alpha_1\|\bp\|+\alpha_0
\eeq
where $\alpha_0,\alpha_1<1$ are constants whose values can be selected
empirically.

In terms of \eqref{StepLEv1}, condition \eqref{bhalpha} reads
\beq  \label{bg}
  \|\bD_{\lambda}^{-1}\bg\| \leq h_{\max},
  \qquad \bg=\bQ^T\nabla\mathcal{F}
\eeq
For simplicity we replace the 2-norm with the maximum norm and get
\beq
    \left|\frac{\bg_1}{d_1+\lambda}\right|\leq h_{\max}
    \quad\text{and}\quad
    \left|\frac{\bg_2}{d_2+\lambda}\right|\leq h_{\max}
\eeq
which gives a lower bound on $\lambda$:
\begin{equation}\label{stepadjutment}
    \lambda\geq \lambda_{\min}
    =\max\left\{\frac{|\bg_1|}{h_{\max}}-d_1,\frac{|\bg_2|}{h_{\max}}-d_2\right\}
\end{equation}
This not only enforces \eqref{bhalpha}, but also guarantees that
$\cH_{\lambda}$ is positive definite. Rather than imposing
\eqref{stepadjutment}, we adjust $\lambda$ as usual: if the iteration
is accepted, $\lambda$ decreases by a certain factor ($\sim 0.1$),
otherwise it increases by a certain factor ($\sim 10$). After the AR2
rule is adopted (i.e., once we get $\|\nabla\cF\|<\varepsilon_\ast$),
we actually set $\lambda=0$ after every successful iteration, so that
the convergence would be truly quadratic.

To summarize, here is the scheme of one iteration of our algorithm:\\

\noindent\textbf{One iteration of our algorithm} \begin{itemize}
  \item {Given $\bp=(a,b)$, compute $\cF(\bp)$, $\nabla \cF(\bp)$, and $\cH(\bp)$}
  \item \textbf{if} $\|\nabla \mathcal{F}\|<\varepsilon_{\ast}$ \textbf{then}
  switch from \textbf{AR1} rule to \textbf{AR2} rule
  \item Compute the eigendecomposition \eqref{eigenH} of the matrix $\cH(\bp)$
  \item Compute the vector $\bg=(\bg_1,\bg_2)$ by \eqref{bg}
  \item Compute $h_{\max}$ by \eqref{bhalpha} and $\lambda_{\min}$ by
  \eqref{stepadjutment}
   \item If \textbf{AR2} rule applies, set $\lambda=0$
  \item \textbf{while}
  \begin{itemize}
   \item[$\circ$] If $\lambda<\lambda_{\min}$, set
       $\lambda=\lambda_{\min}$
   \item[$\circ$] Compute the step $\bh$ by \eqref{StepLEv1}
    \item[$\circ$] \textbf{if} $\|\textbf{h}\|<\beps\|\bp\|$
        \textbf{then} terminate iterations and EXIT
       \item[$\circ$] Compute $\bp'=\bp+\bh$ and $\cF(\bp')$ with
           $\nabla \cF(\bp')$
    \item[$\circ$] \textbf{if AR1} applies:
    \begin{itemize}
      \item \textbf{if} $\mathcal{F}(\bp')<\mathcal{F}(\bp)$
          \textbf{then} set $\bp_{\rm next}=\bp'$, reduce
          $\lambda$, \textbf{break}
    \item \textbf{else} (reject $\bp'$) increase $\lambda$,
        \textbf{continue}
    \end{itemize}
    \item[$\circ$] \textbf{if AR2} applies:
    \begin{itemize}
      \item \textbf{if}
          $\|\nabla\cF(\bp')\|<\|\nabla\cF(\bp)\|$\\
      \textbf{then} set $\bp_{\rm next}=\bp'$, reduce
          $\lambda$, \textbf{break}  \item\textbf{else} (reject
          $\bp'$) increase $\lambda$,
    \textbf{continue}
    \end{itemize}
\end{itemize}
  \item \textbf{end while}
\end{itemize}

\section{Big Circle Formulas}\label{NewFormulas}

\begin{figure}[htb]
\begin{center}
  \includegraphics[width=6in]{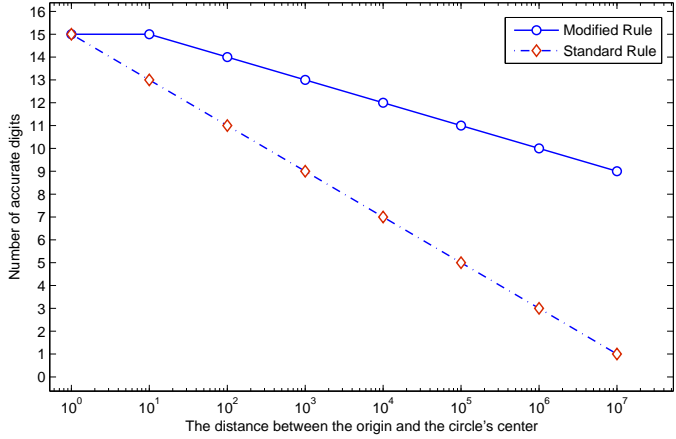}
  \caption{The number of accurate digits in the value of $\cF$ computed by the standard
  rule \eqref{F1a} and by the modified rule \eqref{cFBC}. The picture is almost identical
  for the gradient and Hessian.}\label{FF}
\end{center}
\end{figure}

If our fitting circle is large and still passes near the data points,
then either $a$ or $b$ must be large. In that case we use modified
formulas for $\cF$ and its derivatives to avoid catastrophic
cancelations and minimize round off errors. We use polar coordinates
$D^2=a^2+b^2$ and $\theta\in [0,2\pi)$ so that $a=D\cos \theta$ and
$b=D\sin\theta$. We denote $\delta=1/D$ and $z_i = x_i^2+y_i^2$. Now
$$
   r_i=\sqrt{(a-x_i)^2+(b-y_i)^2} = Dw_i
$$
where $w_i=\sqrt{1-2\delta p_i+\delta^2z_i}$ and $p_i=x_i\cos\theta +
y_i\sin\theta$. We also have $r_i = D+\gamma_i$ where
$$
  \gamma_i = D(w_i-1) = -\frac{\tau_i}{1+w_i},
  \qquad \tau_i = 2p_i-\delta z_i
$$
Further modification gives
$$
  \gamma_i =-p_i+\delta g_i,
  \qquad g_i = \frac{z_i+p_i\gamma_i}{2+\delta\gamma_i}
$$
Averaging over $i=1,\ldots,n$ gives $\overline{p}=0$ (because the data
is centered so that $\overline{x} = \overline{y}=0$) and $\overline{r}
= D+\delta \overline{g}$, thus \eqref{F1a} becomes
\beq  \label{cFBC}
  \cF = -2\overline{g}-\delta^2 \overline{g}^2
\eeq
We found that \eqref{cFBC} gives a more accurate value of $\cF$ than
\eqref{F1a} even for relatively small circles, such as $1<D<10$. For
larger circles, $D\geq 10$, \eqref{cFBC} remains numerically stable
while \eqref{F1a} breaks down (see Figure~\ref{FF}).

Similar modifications can be made for the formulas \eqref{Gradient} and
\eqref{Hessian}. We omit algebraic details and give the final results.
Denote
$$
  \alpha_i = (x_i+\gamma_i\cos\theta)/w_i,\qquad
  \beta_i = (y_i+\gamma_i\sin\theta)/w_i
$$
$$
  \eta_i = \frac{1}{1+\delta\gamma_i},\qquad
  \kappa_i = \frac{\gamma_i}{2+\delta\gamma_i}
$$
Then using our sample mean notation we define
$$
   P = \tfrac 12\,(\overline{\tau\gamma\eta}-\delta\, \overline{\tau\gamma\eta\kappa}),
   \qquad Q=\tfrac 12\,(\overline{\tau\kappa}+\overline{z})
$$
and
$$
  X = \overline{x\gamma\eta},\qquad  Y = \overline{y\gamma\eta}
$$
Now the gradient is given by
\begin{equation}\label{GradientBC}
    \tfrac{1}{2}\,\nabla\mathcal{F}=\delta\left[
                           \begin{array}{c}
                             A\cos\theta-BX \\
                             A\sin\theta-BY \\
                           \end{array}
                         \right]
\end{equation}
where
$$
  A=P+\delta^2(P+Q)Q,\qquad
  B=1+\delta^2 Q
$$
The Hessian is
\begin{equation}\label{HessianBC}
    \tfrac{1}{2}\,\mathcal{H}=\delta^2\left[
                            \begin{array}{cc}
                               U(2\bc-\delta^2U)-Q\bs^2-BN &
                               U\bs+V\bc-\delta^2UV+Q\bs\bc+BL \\
                              U\bs+V\bc-\delta^2UV+Q\bs\bc+BL &
                              V(2\bs-\delta^2V)-Q\bc^2-BM \\
                            \end{array}
                          \right]
\end{equation}
where we use shorthand notation $\bc = \cos\theta$ and $\bs=\sin\theta$
and denote
$$
  U=(P+Q)\bc-X,\qquad V=(P+Q)\bs-Y
$$
and
\begin{align*}
  M&=(\overline{\gamma\gamma\eta}-Q)\bc^2
  +2(\overline{\alpha\gamma\eta}-U)\bc
  +\overline{\alpha\alpha\eta}\\
  N&=(\overline{\gamma\gamma\eta}-Q)\bs^2
  +2(\overline{\beta\gamma\eta}-V)\bs
  +\overline{\beta\beta\eta}\\
  L&=(\overline{\gamma\gamma\eta}-Q)\bc\bs
  +(\overline{\alpha\gamma\eta}-U)\bs
  +(\overline{\beta\gamma\eta}-V)\bc
  +\overline{\alpha\beta\eta}
\end{align*}
The above formulas look more complicated than
\eqref{Gradient}--\eqref{Hessian}. Indeed a careful flop count shows
that they require $55n+$const flops, while
\eqref{Gradient}--\eqref{Hessian} require $19n+$const flops, so the
computational time roughly triples.

\section{Preventing divergence}\label{escape valley}

It is shown in \cite{Cbook} that the graph of the objective function
$\cF(a,b)$ has two valleys stretching toward infinity in opposite
directions. In one valley $\cF(a,b)$ decreases toward its global
minimum, so all the standard algorithms tend to move toward the minimum
and converge. The other valley is separated from the minimum by a
ridge, and in that valley $\cF(a,b)$ decreases as the point $(a,b)$
moves along the valley bottom toward infinity. Thus standard algorithms
tend to diverge. We refer to \cite[Sec.~3.7]{Cbook} for a detailed
account.

To prevent divergence, our program detects whether the current
iteration falls in the ``wrong'' valley, in which case the algorithm
restarts from a point in the ``right'' valley.

The valleys stretch along the eigenvector corresponding to the smaller
eigenvalue of the scatter matrix. The latter is given by
\begin{equation}\label{scatter}
    \bS=\left[
        \begin{array}{cc}
         \overline{xx} & \overline{xy} \\
         \overline{xy} & \overline{yy} \\
        \end{array}
      \right]
\end{equation}
(remember we assumed $\overline{x} = \overline{y} = 0$). Now suppose
the coordinate system is rotated so that the $x$ direction is aligned
with the major eigenvector of $\bS$ and the $y$ direction with its
minor eigenvector. This gives $\overline{xy}=0$ and $\overline{xx} >
\overline{yy}$. Now the valleys stretch in the $y$ (vertical)
direction.

Under these conditions it was proven in \cite[Sec.~3.9]{Cbook} that the
location of the ``wrong'' valley is determined by the sign of
$\overline{xxy} = \frac 1n\sum x_i^2y_i$ as follows:
\begin{itemize}
\item[(a)] if $\overline{xxy} >0$, then the wrong valley lies
    \emph{below} the $x$ axis;
\item[(b)] if $\overline{xxy} <0$, then the wrong valley lies
    \emph{above} the $x$ axis.
\end{itemize}
Our algorithm includes the following block preventing divergence in the
wrong valley (here $L>0$ is a large constant; we use $L=100$):\\

\noindent\textbf{Divergence prevention} \begin{itemize}
  \item \textbf{if} $|a|>L$ or $|b|>L$ \textbf{then}
  \item \textbf{if} the above condition occurs for the first time \textbf{then}
  \begin{itemize}
  \item[$\circ$] Compute the components of the scatter matrix $\bS$
  \item[$\circ$] Compute an eigendecomposition of $\bS$
  \item[$\circ$] Rotate the coordinate system aligning the $x$
      direction with the major eigenvector of $\bS$
  \item[$\circ$] Compute $\overline{xxy}$ and record its sign,
      $Z=$sign$(\overline{xxy})$
  \end{itemize}
  \item Compute the center coordinates $(a,b)$ in the rotated
      system
  \item \textbf{if} $Zb<0$ (i.e., if the center is in the ``wrong'' valley) \textbf{then}
  \item Restart the algorithm from a point in the ``right'' valley
  \end{itemize}

As a restarting point we choose $(0,ZL)$ and rotate it back to the
original coordinate system.

Lastly, the best fitting circle may not exist at all
\cite{Cbook,Ni02,ZC06a}. In that case the data points are best fitted
by a line (more precisely, by the line passing through the origin and
spanned by the major eigenvector of $\bS$). It was proven in
\cite[Sec.~3.9]{Cbook} that this event can only occur if
$\overline{xxy}=0$. Thus our algorithm abandons the search for the best
circle whenever $\max\{|a|,|b|\}>L$ and $|\overline{xxy}|=O(\beps)$. In
that case it returns the best fitting line.

\section{Numerical Results}\label{Numerical results}

We have compared the proposed algorithm with a few others in a series
of computer tests. As competitors, we chose the Gauss-Newton (GN)
method described in \cite{GGS94} (the code was downloaded from the
author's web page), the Levenberg-Marquardt (LM) method (see
\cite[Sec.~4.7]{Cbook}), and the Chernov-Lesort (CL) fit based on
algebraic circle parameters (see \cite{CL1}).

All these fits use the standard stopping rule: they terminate
iterations whenever the step gets smaller than $\beps^{1/2}$, thus they
only achieve suboptimal accuracy $\cO(\beps^{1/2})$. The GN method is
actually able to achieve the optimal accuracy $\cO(\beps)$ if it
continues iterations until the step gets smaller than $\beps$. We
included this modified version (GNm) in our tests.

In our tests we have generated samples of $n=8$ points randomly, with a
uniform distribution in the square $[-1,1] \times [-1,1]$. Each sample
was then centered and scaled as described in Section~\ref{Reducing}.
Uniform distribution produced totally ``chaotic'' samples without any
predefined pattern. It is, in a sense, the worst case scenario,
corresponding to very large noise in data. Thus our algorithms were
tested under the most challenging conditions.

We initialized our fitting algorithms in two different ways. First, we
applied a non-iterative algebraic circle fit, such as K{\aa}sa fit
\cite{Ka76}. Second, we chose the center $(a,b)$ randomly in the square
$[-1,1] \times [-1,1]$ and computed the radius $R$ by \eqref{R}.

Table~\ref{DIVtable} shows how frequently each algorithm diverges and
how many iterations it takes to converge (whenever they do converge).
The GN and LM mostly diverge when the iteration gets into the ``wrong
valley''. The CL fit is designed to converge all the time, but due to
its suboptimal accuracy it occasionally stalls. Our results are
consistent with the ones reported in \cite{CL1}.

\begin{table}[h]
{\footnotesize
\begin{center}
\begin{tabular}{l|c|c|c}
                  \multicolumn{3}{c} {Initial guess} & \\
                  \cline{2-3}
                  & Algebraic   & Randomly chosen & Average \# of\\
                  &  K{\aa}sa fit       & in $[-5,5]\times [-5,5]$ & iterations \\ \hline
  New             & 0\%         & 0\% & 8.1\\
  GNm             & 0.07\%      & 75\% & 29.7 \\
  GN             & 0.07\%      & 75\% & 11.4\\
  LM             & 0.07\%      & 23.5\% & 11.8\\
  CL              & 0\%        & 0.02\% & 11.7\\
  \hline
\end{tabular}
\caption{Percentage of divergencies and the average number of iterations}
\label{DIVtable}
\end{center}}
\end{table}

Next we tested the accuracy of each algorithm. For each generated
sample we first fitted a very precise circle by using the built-in
function \verb"vpa" (available in MATLAB Symbolic Toolbox) that is able
to keep track of 32 (and more) accurate decimal digits. We regard it as
``ideal'' circle. Now each algorithm produced its own fitting circle
and we computed the relative error $E$ in the estimated circle
parameters, versus the ``ideal'' circle. Then we found $k=[-\log_{10}
E]$, which is the number of zeros after the decimal point in the
digital representation of $E$. This can be roughly interpreted as the
number of correct (trustworthy) decimal digits in the parameters of the
fitted circle. Whenever $k \geq 15$, the approximation reaches the
desired ``superaccuracy''. The cases $k \le 14$ are regarded as ``bad
enough'' to be of interest to us and are recorded.

Table~\ref{TableErrors} shows the number of recorded ``bad'' cases for
each $k=1,2,...,14$ for each algorithm, after $10^4$ runs. The table
does not include the super accurate results ($k\geq 15$). All the
algorithms here were initialized by the K{\aa}sa algebraic circle fit.
We only included runs where all the algorithms converged to the same
minimum of the objective function $\cF$ (in all these cases
$E<10^{-2}$).

\begin{table}
{\footnotesize
\begin{center}
\begin{tabular}{c| crrrrrrrrrrr}
        &1 to 3& 4  & 5   & 6    & 7    & 8    & 9   & 10 & 11 & 12& 13  & 14\\ \hline
  New   &0     & 0  & 0   & 0    & 0    & 0    & 0   & 0  & 2  & 2 & 11  & 31   \\ 
  GNm  &0     & 2   & 1   &  0   & 2    & 6    & 2   & 3  & 2  & 8 & 13  & 69  \\ 
  GN   &0     & 0   & 59  & 2459 & 6215 & 1210 & 55  & 2  & 0  & 0 & 0   & 0 \\ 
  LM   &0     & 2   & 174 & 8427 & 1390 & 7    & 0   & 0  & 0  & 0 & 0   & 0  \\ 
  CL    &1     & 1  & 96  & 3270 & 5650 & 953  & 28  & 1  & 0  & 0 & 0   & 0 \\
  \hline
\end{tabular}
\vspace{0.3cm}\caption{The distribution of ``bad'' cases over
$k=1,\ldots,14$. The total number of runs is $10^4$. Superaccurate
results ($k\geq 15$) are not shown. } \label{TableErrors}
\end{center}}
\end{table}

The table shows that the methods GN, LM, CL with a suboptimal stopping
rule (terminating iterations once the step gets smaller than
$\beps^{1/2}$) give suboptimal accuracy with $5 \leq k \leq 9$ in most
cases. The modified GN and our new method use the optimal stopping rule
and achieve better accuracy $k\geq 14$ in most cases. But the modified
GN falters occasionally having rare bad cases down to $k=4$. And it
takes almost 30 iterations, on average to reach the desired accuracy,
while our method takes only 8 iterations.

Our numerical results demonstrate the superiority of our new algorithm
over the main existing algorithms.

\medskip\noindent\textbf{Acknowledgement}.
N.C.\ was partially supported by National Science Foundation, grant
DMS-0969187.


\end{document}